\documentclass[journal,twoside,web]{ieeecolor}
\usepackage{tmi}
\usepackage{cite}
\usepackage{amsmath,amssymb,amsfonts}
\usepackage{algorithmic}
\usepackage[ruled,linesnumbered]{algorithm2e}
\usepackage{multirow}

\usepackage[colorlinks,
            linkcolor=red,
            anchorcolor=red,
            citecolor=green]{hyperref}

\usepackage{graphicx}
\usepackage{textcomp}
\def\BibTeX{{\rm B\kern-.05em{\sc i\kern-.025em b}\kern-.08em
    T\kern-.1667em\lower.7ex\hbox{E}\kern-.125emX}}
\begin{document}
\title{Discrepancy Matters: Learning from Inconsistent Decoder Features for Consistent Semi-supervised Medical Image Segmentation}
\author{Qingjie Zeng, Yutong Xie, Zilin Lu, Mengkang Lu and Yong Xia
\thanks{This work was supported by ********. (Qingjie Zeng and Yutong Xie contributed equally to this work.) (Corresponding author: Yong Xia.)}
\thanks{Qingjie Zeng, Zilin Lu, Mengkang Lu and Yong Xia are with the National Engineering Laboratory for Integrated Aero-Space-Ground-Ocean Big Data Application Technology, School of Computer Science and Engineering,
Northwestern Polytechnical University, Xi’an 710072, China  (e-mail: qjzeng@mail.nwpu.edu.cn; luzl@mail.nwpu.edu.cn; lmk@mail.nwpu.edu.cn; yxia@nwpu.edu.cn)}
\thanks{Yutong Xie is with the Australian Institute for Machine Learning, The University of Adelaide, Adelaide  SA 5000, Australia (e-mail: yutong.xie678@gmail.com).}
}

\maketitle

\begin{abstract}
Semi-supervised learning (SSL) has been proven beneficial for mitigating the issue of limited labeled data especially on the task of volumetric medical image segmentation. Unlike previous SSL methods which focus on exploring highly confident pseudo-labels or developing consistency regularization schemes, our empirical findings suggest that inconsistent decoder features emerge naturally when two decoders strive to generate consistent predictions. Based on the observation, we first analyze the treasure of discrepancy in learning towards consistency, under both pseudo-labeling and consistency regularization settings, and subsequently propose a novel SSL method called LeFeD, which learns the feature-level discrepancy obtained from two decoders, by feeding the discrepancy as a feedback signal to the encoder. The core design of LeFeD is to enlarge the difference by training differentiated decoders, and then learn from the inconsistent information iteratively. We evaluate LeFeD against eight state-of-the-art (SOTA) methods on three public datasets. Experiments show LeFeD surpasses competitors without any bells and whistles such as uncertainty estimation and strong constraints, as well as setting a new state-of-the-art for semi-supervised medical image segmentation. Code is available at \textcolor{cyan}{https://github.com/maxwell0027/LeFeD}
\end{abstract}


\begin{figure}[t]
    \centering
    \includegraphics[width=0.5\textwidth]{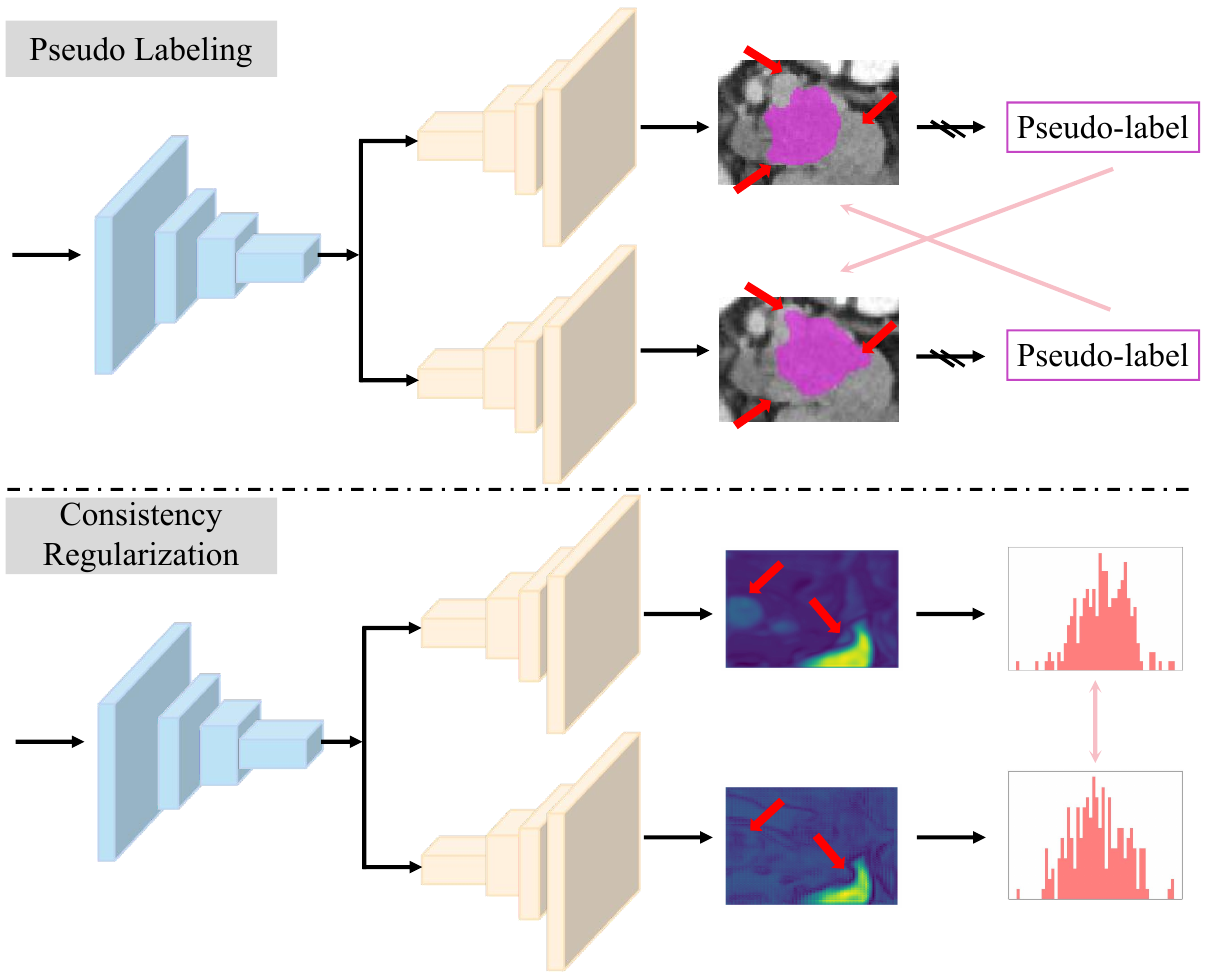}
    \caption{A brief illustration of the significance of inconsistent predictions in semi-supervised learning (SSL). \textbf{Top:} cross-pseudo supervision (pseudo-labeling) and \textbf{Bottom:} consistent logical distribution (consistency regularization). Inconsistent regions are highlighted by \textcolor{red}{red arrow}. SSL can be concluded as learning to be consistent by learning from naturally generated inconsistency.}
    \label{demo}
\end{figure}

\section{Introduction}
\label{sec:introduction}
\IEEEPARstart{A}{ccurate} segmentation of medical images is a crucial task in computer-aided diagnosis~\cite{shen2017deep}. Deep learning models trained on large-scale datasets have recently shown promising performance on this task~\cite{kather2019predicting,krizhevsky2017imagenet}. However, collecting medical image datasets requires ineluctably expertise for data annotation, which is time-consuming and labor-intensive, especially for volumetric data. Considering unlabeled data are relatively easier to collect from clinical sites, semi-supervised learning (SSL)~\cite{cheplygina2019not,chen2022semi} has attracted increasing research attention due to its ability 
to improve model generalization by leveraging massive unlabeled data to augment limited labeled data.

According to the usage of unlabeled data, the paradigm of SSL can be approximately categorized into pseudo-labeling~\cite{liu2022acpl,zhang2022boostmis,rizve2021defense} and consistency regularization~\cite{you2022simcvd,yu2019uncertainty}.
The first category of SSL methods focuses on generating accurate pseudo-labels. For instance, model ensemble was employed in the teacher-student framework to enhance pseudo-label quality~\cite{xiang2022fussnet, bai2023bidirectional}, and various criteria were defined to select accurately pseudo-labeled data ~\cite{zeng2023pefat,lidividemix}. 
The second category of SSL methods put emphasis on designing the regularization that enforces the model to give consistent outputs for an input and its realistically perturbed variants. The consistency regularization can be the constraints imposed at either the data-level~\cite{xie2022IICS,liu2020semi}, task-level~\cite{luo2021semi}, or prediction-level ~\cite{sohn2020fixmatch}.
%
Despite the differences of pseudo-labeling and consistency regularization, they share the same crux that is learning invariant predictions by gradually learning from the inconsistency. For example,~\cite{sohn2020fixmatch} aligns the pseudo-label of strongly-augmented branch to the weakly-augmented branch, and~\cite{luo2022semicnn_trans} keeps the logits distribution similar between predictions of CNN and Transformer.

To better realize this, we present a brief view for the workflow of pseudo-labeling and consistency regularization. As Fig.~\ref{demo} shows, the SSL framework is composed of a single encoder and two decoders – a structure extensively employed in both pseudo-labeling~\cite{yao2022enhancing, chen2021semiCPS} and consistency regularization methods~\cite{wu2021semi, yang2023revisiting}.
 Let us consider an instance where cross-pseudo supervision (a pseudo-labeling strategy displayed in the top of Fig.~\ref{demo}) is utilized. In this scenario, one decoder's pseudo-label is used to oversee the predictions of the other. It is in this context that inconsistent predictions become significant as they can provide complementary information.
Similarly, if we maintain the logical distribution similar for learning from unlabeled data (for example, using KL divergence – a common consistency-based strategy exhibited in the bottom of Fig.~\ref{demo}) between both branches, inconsistent predictions retain a crucial function. This is because the gradient primarily originates from the losses computed within these areas.
From these observations, it becomes evident that inconsistency plays a pivotal role in promoting consistency in learning. Although prior SSL methods have effectively leveraged unlabeled data from the perspective of consistent learning, they have overlooked the natural emergence of inconsistent information when decoders attempt to produce inherently consistent predictions. Moreover, they have failed to acknowledge the significance of discrepancies between those two decoders.


To this end, we propose a novel SSL method called \textbf{Le}arning From the \textbf{Fe}ature-level \textbf{D}iscrepancy (LeFeD) from the perspective of learning inconsistent decoder features.
Our hypothesis is that these discrepancies play a significant role in consistency learning, and properly harnessing this inconsistent information can enhance model performance. Our strategy distinguishes itself from existing methods on two fronts. Firstly, instead of primarily focusing on creating constraints to ensure prediction consistency, we place emphasis on feature discrepancy. Secondly, rather than striving to improve pseudo-label quality, we leverage the discrepancies to augment learning.
In implementation, we first try to enlarge the discrepancy by training two differentiated decoders using distinct loss functions and deep supervision, and then iteratively learn from the inconsistency obtained at all scales. 
Our main contributions are three-fold.
\begin{itemize}
\item[$\bullet$] We propose a novel perspective for SSL, $i.e.$, learning from the inconsistent features produced by two differentiated decoders.
\item[$\bullet$] We observe the phenomenon that, when two decoders attempt to make consistent predictions, there always exists a discrepancy between two predictions, whose contribution to model performance has been verified empirically.
\item[$\bullet$] We propose an accurate SSL method called LeFeD, which beats eight advanced SSL methods on three public medical image datasets, setting a new state of the art for semi-supervised medical image segmentation.
\end{itemize}

\begin{figure*}[t]
    \centering
    \includegraphics[width=0.9\textwidth]{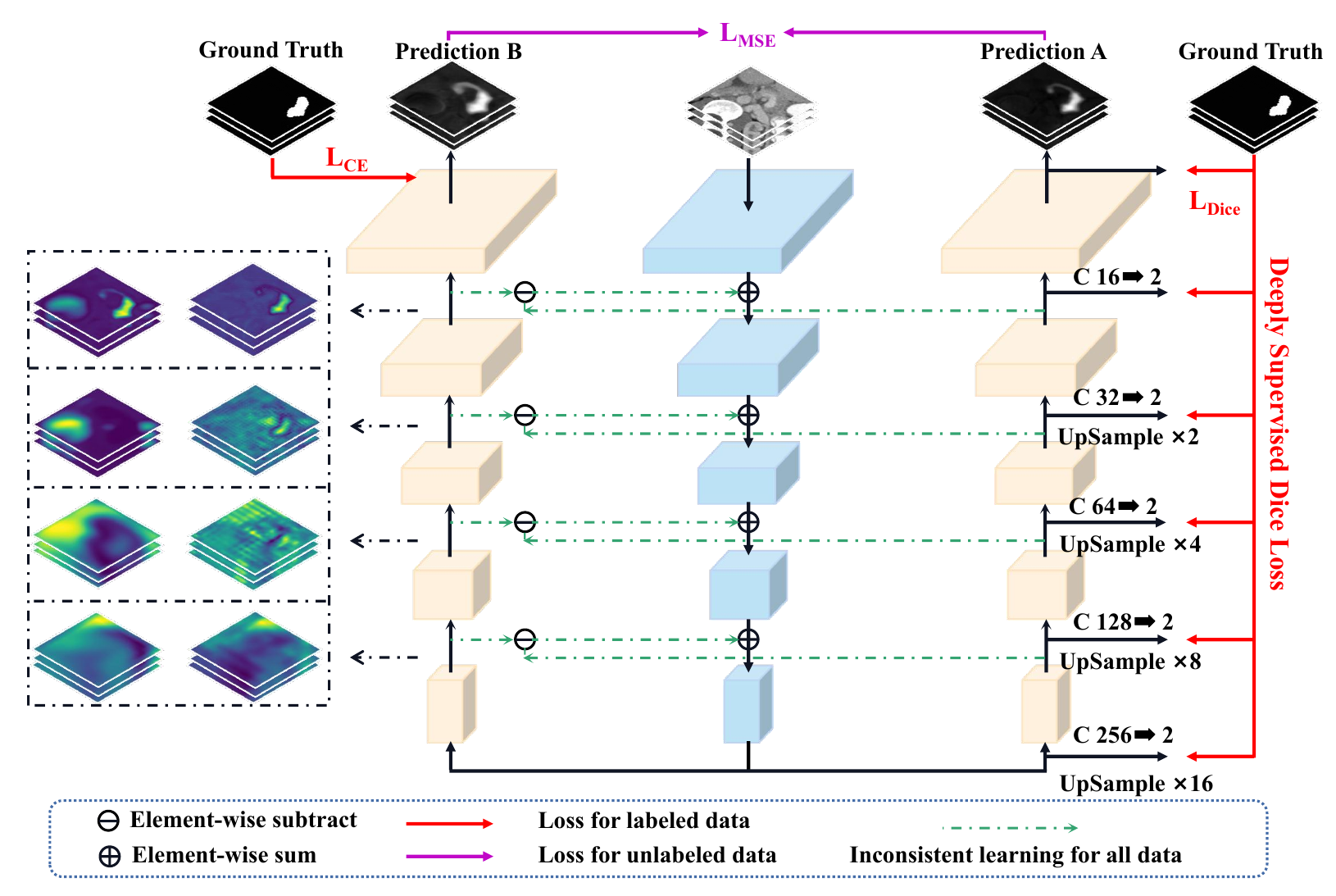}
    \caption{Illustration of our LeFeD model. The paired feature maps on the left were produced by the two decoders and were presented as examples to show the discrepancies. We can find decoder features vary significantly in all scales, and LeFeD is designed to learn from these discrepancies by feeding the discrepancies as supplementary information to the encoder.}
    \label{framework}
\end{figure*}

\section{Related Works}

\subsection{Medical Image Segmentation}
It is well-acknowledged that delineating voxel-wise medical images is expensive and tedious, as well as requiring clinical expertise. Recently, deep learning methods present superior performance in automatic medical image segmentation, and a series of frameworks have emerged. For the CNN-based network, nnU-Net~\cite{isensee2021nnu} extends U-Net~\cite{ronneberger2015unet} and validates the necessity of image pre-processing/post-processing and abundant data augmentation. With the rise of ViT~\cite{ViT2020}, many attempts are made to combine the advantages of Transformer and CNN, including nnFormer~\cite{zhou2021nnformer}, TransUNet~\cite{chen2021transunet}, CoTr~\cite{xie2021cotr}, Swin-Unet~\cite{cao2022swinunet}, $etc$. These methods use convolutions and long-range self-attention to capture both local and global features, and present good results on numerous medical image segmentation tasks.

Meanwhile, there are different settings in medical image segmentation, containing self-supervised learning~\cite{jing2020self}, few-shot learning~\cite{feng2021interactive}, weakly-supervised~\cite{hassan2022supervised} and semi-supervised learning~\cite{cheplygina2019not,chen2022semi}. In this paper, we focus on the topic of semi-supervised learning, which aims to achieve supervised performance using only limited labeled data along with substantial unlabeled data, thereby mitigating the cost of data annotation.

\subsection{Semi-supervised Learning in Medical Image Analysis}
Among existing SSL methods, pseudo-labeling and consistency regularization are the most popular. Pseudo-labeling intends to discover high-quality pseudo-labels for re-training. For instance, 
BoostMIS~\cite{zhang2022boostmis} takes into account the model performance in different training stages and uses an adaptive threshold to select pseudo-labels.
UA-MT~\cite{yu2019uncertainty} treats the prediction made by the teacher branch as the pseudo-label and uses it to supervise the student branch. 
ACPL~\cite{liu2022acpl} introduces an anti-curriculum learning scheme, which trains models using hard samples first, and then easy samples. 
PEFAT~\cite{zeng2023pefat} selects trustworthy pseudo-labels from the perspective of loss distribution.
As for the consistency-based SSL methods, different types of constraints are commonly investigated. For example, 
I$^{2}$CS~\cite{xie2022IICS} simulates the relation between labeled and unlabeled data by calculating attention maps. 
URPC~\cite{luo2022semi_urpc} designs uncertainty rectified pyramid consistency for multi-scale features obtained from one decoder.
DTC~\cite{luo2021semi} discovers the consistency between segmentation and regression tasks. 
MC-Net+~\cite{wu2022mutual} uses three decoders with different structures and enforces the outputs to be consistent.

By contrast, our LeFeD neither focuses on pseudo-label generation nor separately processing the regions with different confidence (or uncertainty) levels. It pays attention to the discrepancy produced in learning towards consistency.

\subsection{Inconsistent Learning in Medical Image Analysis}
We also review the application of inconsistent learning in medical image processing. In adversarial training, VANT-GAN~\cite{zia2022vant} formulates visual attribution according to the discrepancy map between normal and abnormal counterparts to learn abnormal-to-normal mapping. GAN-BiLSTM-CRF~\cite{yu2020adversarial} incorporates a generator and a discriminator to address the issue of annotation inconsistency for those pseudo-labels generated by active learning. Specifically, in semi-supervised learning, CoraNet~\cite{shi2021inconsistency} treats the consistent and inconsistent predicted regions separately and designs an uncertainty estimation strategy to filter out the inconsistent pseudo-labels. NonAdjLoss~\cite{ganaye2019removing} proposes a non-adjacency constraint to discover segmentation anomalies, which also directly removes inconsistent segmentation. All these methods do not realise the significance of inconsistent information in unlabeled data mining, and most of them tend to discard the inherent discrepancy. 

By comparison, we analyze the importance of discrepancy in detail from the perspective both of pseudo-labeling and consistency regularization. To our knowledge, we are the first to advocate learning from inconsistency under the SSL setting.

\section{Method}

As delineated in Section~\ref{sec:introduction}, we posit that the observed discrepancies play a pivotal role, particularly in the context of achieving learning consistency. Such nuances, when harnessed appropriately, can be invaluable. 
As illustrated in Fig.~\ref{framework}, the foundational concept of LeFeD is anchored on two key steps: firstly, amplifying the variance between the feature maps generated by dual decoders at each hierarchical level, and subsequently, assimilating insights from this inconsistency in a recursive manner.
To engender this diversified discrepancy, we approach it from two strategic dimensions.
First, distinct loss functions are designated for each of the decoders. The cross-entropy (CE) loss is tailored to foster precise voxel-wise predictions, juxtaposed with the Dice loss which is oriented towards emphasizing region-wise predictions.
Second, deep supervision~\cite{lee2015deeply} is employed to enlarge the difference. 
Furthermore, the discrepancy obtained from the last iteration is integrated into the encoder in the current iteration, serving as auxiliary information. 
It means that each sample will be learned several times with the discrepancy except the first time in each iteration.

\begin{table*}[]
\caption{Performance comparison with other eight state-of-the-art methods on the pancreas dataset, in the scenario of leveraging 10\% and 20\% labeled data. Improvements compared with BCP~\cite{bai2023bidirectional} are \textcolor{blue}{highlighted}. }\label{table1}

\centering 
\begin{tabular}{c|cccc|cccc}
\hline
\multirow{2}{*}{Method} 
& \multicolumn{4}{c|}{Pancreas (10\%/6 labeled data)}          
& \multicolumn{4}{c}{Pancreas (20\%/12 labeled data)}      \\\cline{2-9} 
& \multicolumn{1}{c|}{Dice~$\uparrow$} & \multicolumn{1}{c|}{Jaccard~$\uparrow$} & \multicolumn{1}{c|}{ASD~$\downarrow$} & 95HD~$\downarrow$ & \multicolumn{1}{c|}{Dice~$\uparrow$} & \multicolumn{1}{c|}{Jaccard~$\uparrow$} & \multicolumn{1}{c|}{ASD~$\downarrow$} & 95HD~$\downarrow$ \\ \hline

VNet& \multicolumn{1}{c|}{55.60}     & \multicolumn{1}{c|}{41.74}        & \multicolumn{1}{c|}{18.63}    & 45.33     & \multicolumn{1}{c|}{72.38}     & \multicolumn{1}{c|}{58.26}    & \multicolumn{1}{c|}{5.89}    & 19.35 \\ \hline\hline

UA-MT~(MICCAI'19)& \multicolumn{1}{c|}{66.34}     & \multicolumn{1}{c|}{53.21}        & \multicolumn{1}{c|}{4.57}    & 17.21     & \multicolumn{1}{c|}{76.10}     & \multicolumn{1}{c|}{62.62}     & \multicolumn{1}{c|}{2.43}   & 10.84     \\ \hline
    
SASSNet~(MICCAI'20)& \multicolumn{1}{c|}{68.78}     & \multicolumn{1}{c|}{53.86}        & \multicolumn{1}{c|}{6.26}    &  19.02    & \multicolumn{1}{c|}{77.66}     & \multicolumn{1}{c|}{64.08}  & \multicolumn{1}{c|}{3.05}    &  10.93    \\ \hline
    
DTC~(AAAI'21)& \multicolumn{1}{c|}{69.21}     & \multicolumn{1}{c|}{54.06}        & \multicolumn{1}{c|}{5.95}    &  17.21    & \multicolumn{1}{c|}{78.27}     & \multicolumn{1}{c|}{64.75}  & \multicolumn{1}{c|}{2.25}    &  8.36   \\ \hline
    
ASE-Net~(TMI'22)& \multicolumn{1}{c|}{71.54}     & \multicolumn{1}{c|}{56.82}        & \multicolumn{1}{c|}{5.73}    &  16.33    & \multicolumn{1}{c|}{79.03}     & \multicolumn{1}{c|}{66.57}  & \multicolumn{1}{c|}{2.30}    &  8.62   \\ \hline

SS-Net~(MICCAI'22)& \multicolumn{1}{c|}{71.76}     & \multicolumn{1}{c|}{57.05}        & \multicolumn{1}{c|}{5.77}    & 17.56     & \multicolumn{1}{c|}{78.98}     & \multicolumn{1}{c|}{66.32}     & \multicolumn{1}{c|}{2.01}   & 8.86     \\ \hline
    
MC-Net+~(MedIA'22)& \multicolumn{1}{c|}{70.00}     & \multicolumn{1}{c|}{55.66}        & \multicolumn{1}{c|}{3.87}    &  16.03    & \multicolumn{1}{c|}{79.37}     & \multicolumn{1}{c|}{66.83}  & \multicolumn{1}{c|}{1.72}    &  8.52    \\ \hline
    
FUSSNet~(MICCAI'22)& \multicolumn{1}{c|}{72.03}     & \multicolumn{1}{c|}{57.82}        & \multicolumn{1}{c|}{5.33}    &  16.77    & \multicolumn{1}{c|}{81.82}     & \multicolumn{1}{c|}{69.76}  & \multicolumn{1}{c|}{1.51}    &  5.42    \\ \hline

BCP~(CVPR'23)& \multicolumn{1}{c|}{73.83}     & \multicolumn{1}{c|}{59.24}        & \multicolumn{1}{c|}{3.72}    &  12.71    & \multicolumn{1}{c|}{\textbf{82.91}}     & \multicolumn{1}{c|}{70.97}  & \multicolumn{1}{c|}{2.25}    &  6.43   \\ \hline\hline
    
LeFeD(Ours)& \multicolumn{1}{c|}{\textbf{75.51}$_{\textcolor{blue}{+1.68}}$}     & \multicolumn{1}{c|}{\textbf{61.51}$_{\textcolor{blue}{+2.27}}$}        & \multicolumn{1}{c|}{\textbf{3.44}$_{\textcolor{blue}{+0.28}}$}    &  \textbf{11.79}$_{\textcolor{blue}{+0.92}}$   & \multicolumn{1}{c|}{82.69$_{\textcolor{red}{-0.22}}$}     & \multicolumn{1}{c|}{\textbf{71.03}$_{\textcolor{blue}{+0.06}}$}  & \multicolumn{1}{c|}{\textbf{1.32}$_{\textcolor{blue}{+0.93}}$}    &  \textbf{5.03$_{\textcolor{blue}{+1.43}}$}    \\ \hline

Fully Supervised& \multicolumn{1}{c|}{83.76}     & \multicolumn{1}{c|}{72.39}        & \multicolumn{1}{c|}{1.01}    &  4.36    & \multicolumn{1}{c|}{83.76}     & \multicolumn{1}{c|}{72.39}  & \multicolumn{1}{c|}{1.01}    &  4.36   \\ \hline\hline
    
\end{tabular}
\end{table*}

\begin{figure*}[t]
    \centering
    \includegraphics[width=0.9\textwidth]{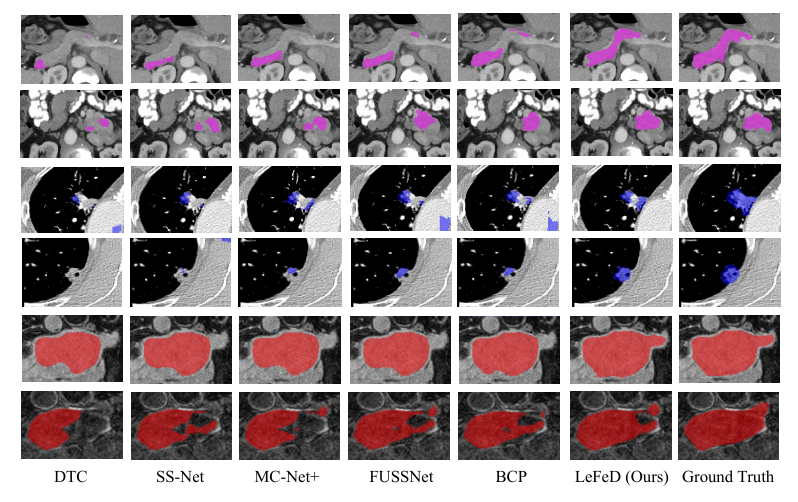}
    \caption{2D visualization results of different SSL methods when training with 10\% labeled data. Top two rows are segmentation results on the pancreas dataset, medium two rows are results on the lung tumor dataset, and bottom two rows are results on the left atrium dataset.}
    \label{segmentation_results}
\end{figure*}

\subsection{Training Differentiated Decoders}

Drawing from intuitive understanding, the greater the variation between decoders—whether in terms of their architectural designs, training methodologies, or input types—the more unique their respective outputs tend to be. 
To effectively harness this inconsistency, we initially employ two decoders distinguished by their up-sampling techniques: one utilizes tri-linear interpolation, while the other employs transposed convolution.
Furthermore, to augment the variance between the decoders, deep supervision is applied exclusively to one of them, ensuring that features across all scales manifest differently. 
In the context of the deeply supervised decoder, we use one convolution layer with the kernel size of 1$\times$1$\times$1 to adjust the output channel. Subsequently, the feature map undergoes up-sampling to align with the dimensions of the ground truth.
Moreover, it's essential to note that while both decoders target identical outcomes on labeled data, their optimization processes are deliberately varied. Specifically, the CE loss is paired with one, while the Dice loss is reserved for the other, ensuring distinctive optimization trajectories for each.
To this end, the objective for model training on labeled data can be formulated as
\begin{equation}
L_{sup} = L_{ds} + L_{dice}(f_{\theta_{A}}(x_{i}), y_{i}) + L_{ce}(f_{\theta_{B}}(x_{i}), y_{i})  
\label{eq1}
\end{equation}

\begin{equation}
L_{ds} = \sum_{s=1}^{S} w_{s} L_{dice}(\zeta(f_{\theta_{A}}^{s}(x_{i})), y_{i})
\label{eq2}
\end{equation}
where $f_{\theta_{A}}$ and $f_{\theta_{B}}$ stand for the decoders that are optimized by Dice loss and CE loss, respectively. $L_{ds}$ is the deep supervision loss, $S$ is the number of scales and $w_{s}$ is a constant to balance the weight of loss calculated from the $s$-th scale. $\zeta$ is the combination of convolution and up-sampling. $D_{L} = \left\{(x_{i},y_{i})\right\}_{i=1}^{N}$ is the labeled data set and $N$ is the number of labeled images. In experiments, $w_{s}$ are set to $\left\{0.8, 0.6, 0.4, 0.2, 0.1 \right\}$ for the deeply supervised loss calculated from low-level to high-level features.

As for the unlabeled data, we do not put emphasis on finding truthworthy pseudo-labels or investigating regularized approaches, but simply employ Mean Squared Error (MSE) loss to make predictions consistent, which is written as
\begin{equation}
L_{unsup} = L_{mse}(f_{\theta_{A}}(x_{u}), f_{\theta_{B}}(x_{u}))
\label{eq3}
\end{equation}
where $D_{U} = \left\{x_{u} \right\}_{u=1}^{M}$ is the unlabeled data set and $M$ is the number of unlabeled images. 
The overall loss is the sum of the supervised and unsupervised losses, formulated as
\begin{equation}
L_{total} = L_{sup} + L_{unsup}
\label{eq4}
\end{equation}

By training differentiated decoders, our model can capture more inconsistent information and thereby show better results.

\begin{table*}[t]
\caption{Performance comparison with other eight state-of-the-art methods on the lung tumor dataset, in the scenario of leveraging 20\% and 30\% labeled data. Improvements compared with BCP~\cite{bai2023bidirectional} are \textcolor{blue}{highlighted}. }\label{table_l}

\centering 
\begin{tabular}{c|cccc|cccc}
\hline
\multirow{2}{*}{Method} & \multicolumn{4}{c|}{Lung tumor (20\%/10 labeled data)}                                       & \multicolumn{4}{c}{Lung tumor (30\%/15 labeled data)}                                         \\ \cline{2-9} 
& \multicolumn{1}{c|}{Dice~$\uparrow$} & \multicolumn{1}{c|}{Jaccard~$\uparrow$} & \multicolumn{1}{c|}{ASD~$\downarrow$} & 95HD~$\downarrow$ & \multicolumn{1}{c|}{Dice~$\uparrow$} & \multicolumn{1}{c|}{Jaccard~$\uparrow$} & \multicolumn{1}{c|}{ASD~$\downarrow$} & 95HD~$\downarrow$ \\ \hline

VNet& \multicolumn{1}{c|}{36.89}     & \multicolumn{1}{c|}{26.00}    & \multicolumn{1}{c|}{19.98}    & 30.47 & \multicolumn{1}{c|}{42.90}     & \multicolumn{1}{c|}{30.29}        & \multicolumn{1}{c|}{5.78}    & 13.79     \\ \hline\hline

UA-MT~(MICCAI'19)& \multicolumn{1}{c|}{44.33}     & \multicolumn{1}{c|}{30.80}     & \multicolumn{1}{c|}{10.28}   & 22.82     & \multicolumn{1}{c|}{56.20}     & \multicolumn{1}{c|}{41.83}        & \multicolumn{1}{c|}{5.24}    & 13.63     \\ \hline
    
SASSNet~(MICCAI'20)& \multicolumn{1}{c|}{45.91}     & \multicolumn{1}{c|}{31.47}  & \multicolumn{1}{c|}{8.25}    &  17.25    & \multicolumn{1}{c|}{58.96}     & \multicolumn{1}{c|}{44.05}        & \multicolumn{1}{c|}{2.49}    &  9.50    \\ \hline
    
DTC~(AAAI'21)& \multicolumn{1}{c|}{48.46}     & \multicolumn{1}{c|}{34.49}  & \multicolumn{1}{c|}{7.70}    &  21.22   & \multicolumn{1}{c|}{59.07}     & \multicolumn{1}{c|}{43.45}        & \multicolumn{1}{c|}{2.32}    &  8.48    \\ \hline

SS-Net~(MICCAI'22)& \multicolumn{1}{c|}{49.78}     & \multicolumn{1}{c|}{34.26}  & \multicolumn{1}{c|}{7.70}    &   21.09   & \multicolumn{1}{c|}{58.56}     & \multicolumn{1}{c|}{42.84}        & \multicolumn{1}{c|}{2.64}    &  9.39    \\ \hline
  
FUSSNet~(MICCAI'22)& \multicolumn{1}{c|}{50.10}     & \multicolumn{1}{c|}{35.35}  & \multicolumn{1}{c|}{6.59}    &   19.18   & \multicolumn{1}{c|}{58.97}     & \multicolumn{1}{c|}{44.83}        & \multicolumn{1}{c|}{2.93}    &  9.22    \\ \hline

MC-Net+~(MedIA'22)& \multicolumn{1}{c|}{50.63}     & \multicolumn{1}{c|}{36.00}     & \multicolumn{1}{c|}{3.59}   & 11.52     & \multicolumn{1}{c|}{59.65}     & \multicolumn{1}{c|}{44.30}        & \multicolumn{1}{c|}{2.78}    & 8.89     \\ \hline
    
ASE-Net~(TMI'22)& \multicolumn{1}{c|}{53.15}     & \multicolumn{1}{c|}{38.29}  & \multicolumn{1}{c|}{3.77}    &  12.87    & \multicolumn{1}{c|}{61.06}     & \multicolumn{1}{c|}{44.92}        & \multicolumn{1}{c|}{2.80}    &  9.10    \\ \hline
    
BCP~(CVPR'23)& \multicolumn{1}{c|}{54.63}     & \multicolumn{1}{c|}{38.13}  & \multicolumn{1}{c|}{3.62}    &  11.77    & \multicolumn{1}{c|}{61.58}     & \multicolumn{1}{c|}{46.44}        & \multicolumn{1}{c|}{2.66}    &  9.05    \\ \hline\hline

LeFeD(Ours)& \multicolumn{1}{c|}{\textbf{56.74}$_{\textcolor{blue}{+2.11}}$}     & \multicolumn{1}{c|}{\textbf{41.30}$_{\textcolor{blue}{+3.17}}$}  & \multicolumn{1}{c|}{\textbf{3.49}$_{\textcolor{blue}{+0.13}}$}    &  \textbf{10.60$_{\textcolor{blue}{+1.17}}$}    & \multicolumn{1}{c|}{\textbf{62.34}$_{\textcolor{blue}{+0.76}}$}     & \multicolumn{1}{c|}{\textbf{47.49}$_{\textcolor{blue}{+1.05}}$}        & \multicolumn{1}{c|}{\textbf{2.24}$_{\textcolor{blue}{+0.42}}$}    &  \textbf{8.39}$_{\textcolor{blue}{+0.66}}$   \\ \hline

Fully Supervised& \multicolumn{1}{c|}{64.04}     & \multicolumn{1}{c|}{49.08}        & \multicolumn{1}{c|}{1.98}    &  7.42    & \multicolumn{1}{c|}{64.04}     & \multicolumn{1}{c|}{49.08}  & \multicolumn{1}{c|}{1.98}    &  7.42   \\ \hline\hline
    
\end{tabular}
\end{table*}

\begin{table*}[t]
\caption{Performance comparison with other eight state-of-the-art methods on the left atrium dataset, in the scenario of leveraging 10\% and 20\% labeled data. Improvements compared with BCP~\cite{bai2023bidirectional} are \textcolor{blue}{highlighted}.}\label{table2}
\centering 
\begin{tabular}{c|cccc|cccc}
\hline
\multirow{2}{*}{Method} & \multicolumn{4}{c|}{Left atrium (10\%/8 labeled data)}                                       & \multicolumn{4}{c}{Left atrium (20\%/16 labeled data)}                                         \\ \cline{2-9} 
& \multicolumn{1}{c|}{Dice~$\uparrow$} & \multicolumn{1}{c|}{Jaccard~$\uparrow$} & \multicolumn{1}{c|}{ASD~$\downarrow$} & 95HD~$\downarrow$ & \multicolumn{1}{c|}{Dice~$\uparrow$} & \multicolumn{1}{c|}{Jaccard~$\uparrow$} & \multicolumn{1}{c|}{ASD~$\downarrow$} & 95HD~$\downarrow$ \\ \hline

VNet& \multicolumn{1}{c|}{82.74}     & \multicolumn{1}{c|}{71.72}        & \multicolumn{1}{c|}{3.26}    & 13.35     & \multicolumn{1}{c|}{84.89}     & \multicolumn{1}{c|}{77.32}    & \multicolumn{1}{c|}{2.97}    & 11.60  \\ \hline\hline

UA-MT~(MICCAI'19)& \multicolumn{1}{c|}{86.28}     & \multicolumn{1}{c|}{76.11}        & \multicolumn{1}{c|}{4.63}    & 18.71     & \multicolumn{1}{c|}{88.74}     & \multicolumn{1}{c|}{79.94}     & \multicolumn{1}{c|}{2.32}   & 8.39     \\ \hline
    
SASSNet~(MICCAI'20)& \multicolumn{1}{c|}{87.54}     & \multicolumn{1}{c|}{78.05}        & \multicolumn{1}{c|}{2.59}    &  9.84    & \multicolumn{1}{c|}{89.54}     & \multicolumn{1}{c|}{81.24}  & \multicolumn{1}{c|}{2.20}    & 8.24    \\ \hline
    
DTC~(AAAI'21)& \multicolumn{1}{c|}{87.51}     & \multicolumn{1}{c|}{78.17}        & \multicolumn{1}{c|}{2.36}    &  8.23    & \multicolumn{1}{c|}{89.42}     & \multicolumn{1}{c|}{80.89}  & \multicolumn{1}{c|}{2.10}    &  7.32   \\ \hline
    
ASE-Net~(TMI'22)& \multicolumn{1}{c|}{87.83}     & \multicolumn{1}{c|}{78.45}        & \multicolumn{1}{c|}{2.17}    &  9.86    & \multicolumn{1}{c|}{90.29}     & \multicolumn{1}{c|}{82.76}  & \multicolumn{1}{c|}{1.64}    &  7.18   \\ \hline

SS-Net~(MICCAI'22)& \multicolumn{1}{c|}{88.55}     & \multicolumn{1}{c|}{79.62}        & \multicolumn{1}{c|}{1.90}    & 7.49     & \multicolumn{1}{c|}{90.98}     & \multicolumn{1}{c|}{83.69}     & \multicolumn{1}{c|}{1.74}   & 5.95     \\ \hline
    
MC-Net+~(MedIA'22)& \multicolumn{1}{c|}{88.96}     & \multicolumn{1}{c|}{80.25}        & \multicolumn{1}{c|}{1.86}    &  7.93    & \multicolumn{1}{c|}{91.07}     & \multicolumn{1}{c|}{83.67}  & \multicolumn{1}{c|}{1.67}    &  5.84   \\ \hline
    
FUSSNet~(MICCAI'22)& \multicolumn{1}{c|}{88.78}     & \multicolumn{1}{c|}{80.18}        & \multicolumn{1}{c|}{1.80}    &  7.82    & \multicolumn{1}{c|}{91.13}     & \multicolumn{1}{c|}{83.79}  & \multicolumn{1}{c|}{1.56}    &  \textbf{5.10}    \\ \hline

BCP~(CVPR'23)& \multicolumn{1}{c|}{89.62}     & \multicolumn{1}{c|}{81.31}        & \multicolumn{1}{c|}{1.76}    &  6.81    & \multicolumn{1}{c|}{91.25}     & \multicolumn{1}{c|}{83.85}  & \multicolumn{1}{c|}{1.47}    &  5.96   \\ \hline\hline
    
LeFeD(Ours)& \multicolumn{1}{c|}{\textbf{90.00}$_{\textcolor{blue}{+0.38}}$}     & \multicolumn{1}{c|}{\textbf{81.99}$_{\textcolor{blue}{+0.68}}$}        & \multicolumn{1}{c|}{\textbf{1.67}$_{\textcolor{blue}{+0.09}}$}    &  \textbf{6.78}$_{\textcolor{blue}{+0.03}}$    & \multicolumn{1}{c|}{\textbf{91.44}$_{\textcolor{blue}{+0.19}}$}     & \multicolumn{1}{c|}{\textbf{84.30}$_{\textcolor{blue}{+0.45}}$}  & \multicolumn{1}{c|}{\textbf{1.39}$_{\textcolor{blue}{+0.08}}$}    &  5.87$_{\textcolor{blue}{+0.09}}$  \\ \hline

Fully Supervised& \multicolumn{1}{c|}{91.47}     & \multicolumn{1}{c|}{84.36}        & \multicolumn{1}{c|}{1.51}    &  5.48    & \multicolumn{1}{c|}{91.47}     & \multicolumn{1}{c|}{84.36}  & \multicolumn{1}{c|}{1.51}    &  5.48   \\ \hline\hline
    
\end{tabular}
\end{table*}

\subsection{Learning from Discrepancy}
As illustrated in Fig.~\ref{demo}, we observe inherent inconsistencies in the features when two decoders strive to produce identical predictions. And we argue that this observed discrepancy can serve as valuable supplementary information, enhancing the model's capacity to learn effectively from unlabeled data. 
Consequently, we propose the incorporation of this discrepancy into the encoder. This strategy involves repeatedly utilizing this supplementary information for multiple iterations per sample, thereby enriching the learning process and potentially improving the model's performance.
Concretely, the process can be formulated as:
\begin{equation}
Output^{t} = \begin{cases} E^{t}(x), &\text{if}~ t = 1, \\ E^{t}(x + \lambda(f_{\theta_{A}}^{t-1}(x) - f_{\theta_{B}}^{t-1}(x))),  &\text {if}~t > 1, \end{cases}
\end{equation}
where $Output^{t}$ is the feature embedding of $t$-th iteration. $E$ symbolizes the encoder and $\lambda$ serves as a hyper-parameter designed to regulate the influence of feature discrepancies (here some subscripts, $e.g.$, $i$, $s$ and $u$ are omitted for the sake of clarity and simplicity of explanation).


In summary, LeFeD is centered on the concept of discrepancy learning. This approach is characterized by the training of distinct decoders through the strategic employment of varied loss functions and deep supervision. The primary objective underpinning this strategy is to accentuate and leverage inconsistent information to a significant degree.

\section{Experiments}
\subsection{Datasets}
We employed three publicly available datasets for the evaluation of our method, including the pancreas dataset~\cite{roth2015deeporgan}, the lung tumor dataset~\cite{antonelli2022medical}, and the left atrium dataset~\cite{xiong2021global}.
The pancreas dataset consists of 82 contrast-enhanced abdomen CT scans, which are split into 62 scans for training and 20 scans for testing. 
The lung tumor dataset includes 63 cases, with a split of 50 for training and 13 for testing purposes. 
The left atrium dataset contains 100 gadolinium-enhanced MR images, out of which 80 serve as training images and the rest, 20, are for testing. 
For both the pancreas and left atrium datasets, we strictly adhered to a consistent pre-processing protocol. This involved center cropping with an enlargement margin of $25$ voxels, respacing to attain an isotropic resolution of 1.0mm $\times$ 1.0mm $\times$ 1.0mm, and normalization to achieve zero mean with unit variance. 
To make fair comparisons with~\cite{lei2022semi,li2020shape,luo2021semi,xiang2022fussnet,wu2022exploring,wu2022mutual}, results were detailed with a label percentage of both 10\% and 20\%. 

For the lung tumor dataset, an initial Hounsfield Units (HU) threshold, ranging from -500 to 275, was employed for voxel value clipping. Thereafter, the aforementioned pre-processing strategy was utilized. Given the heightened complexity associated with tumor segmentation compared to organ segmentation, results were presented with 20\% and 30\% labeled data.

\subsection{Implementation Details}
Following previous works~\cite{lei2022semi,li2020shape,luo2021semi,xiang2022fussnet,wu2022exploring,wu2022mutual}, V-Net~\cite{milletari2016v} was set as the baseline for easy implementation and comparison. For model training, we used an SGD optimizer and set the learning rate to 0.01, the weight decay to 0.0001 and the momentum to 0.9. Hyper-parameter of iteration times $t$ and $\lambda$ were set to 3 and 1e-3. The input size was respectively set to 96$\times$96$\times$96, 96$\times$96$\times$96 and 112$\times$112$\times$80 for pancreas dataset, lung tumor dataset and left atrium dataset. The batch size was 4, containing 2 labeled and 2 unlabeled cubic patches. The whole experiments were implemented by Pytorch~\cite{paszke2019pytorch} with one NVIDIA GeForce RTX 3080 Ti GPU. Evaluation metrics of Dice, Jaccard, Average Surface Distance (ASD) and 95\% Hausdorff Distance (95HD) were reported.

\begin{table*}[t]
\caption{Ablation studies that are conducted on the pancreas dataset with 10\% labeled data. ``IL'' means \textbf{I}nconsistent \textbf{L}earning for the feature discrepancy obtained from all scales. ``DL'' means applying \textbf{D}ifferent \textbf{L}oss. ``DS'' stands for \textbf{D}eep \textbf{S}upervision. ``S1'' means learning the feature discrepancy obtained from \textbf{S}cale1.}\label{table3}
\setlength{\tabcolsep}{3.2mm}
\centering 
\begin{tabular}{c|cccc}
\hline
\multirow{2}{*}{Method} & \multicolumn{4}{c}{Metrics}                                                                                      \\ \cline{2-5}  
                        & \multicolumn{1}{c|}{Dice(\%)~$\uparrow$} & \multicolumn{1}{c|}{Jaccard(\%)~$\uparrow$} & \multicolumn{1}{c|}{ASD(voxel)~$\downarrow$} & 95HD(voxel)~$\downarrow$ \\ \hline
Baseline~(VNet)                & \multicolumn{1}{c|}{68.10}    & \multicolumn{1}{c|}{52.51}       & \multicolumn{1}{c|}{7.22}       & 19.92       \\ \hline
+IL                     & \multicolumn{1}{c|}{71.00}    & \multicolumn{1}{c|}{55.96}       & \multicolumn{1}{c|}{4.34}       & 14.53       \\ \hline
+IL+DL                  & \multicolumn{1}{c|}{72.42}    & \multicolumn{1}{c|}{57.43}       & \multicolumn{1}{c|}{3.61}       & 11.85       \\ \hline
+IL+DS                  & \multicolumn{1}{c|}{73.57}    & \multicolumn{1}{c|}{59.09}       & \multicolumn{1}{c|}{3.90}       & 13.95       \\ \hline \hline
+DL+DS+S1               & \multicolumn{1}{c|}{72.51}    & \multicolumn{1}{c|}{57.77}       & \multicolumn{1}{c|}{5.45}       & 18.45       \\ \hline
+DL+DS+S1+S2             & \multicolumn{1}{c|}{73.61}    & \multicolumn{1}{c|}{59.19}       & \multicolumn{1}{c|}{4.93}       &  15.42      \\ \hline
+DL+DS+S1+S2+S3               & \multicolumn{1}{c|}{74.57}    & \multicolumn{1}{c|}{60.18}       & \multicolumn{1}{c|}{3.98}       & 13.27       \\ \hline
+DL+DS+IL              & \multicolumn{1}{c|}{\textbf{75.51}}    & \multicolumn{1}{c|}{\textbf{61.51}}       & \multicolumn{1}{c|}{\textbf{3.44}}       & \textbf{11.79}       \\ \hline
\end{tabular}
\end{table*}

\begin{figure*}[t]
    \centering
    \includegraphics[width=0.72\textwidth]{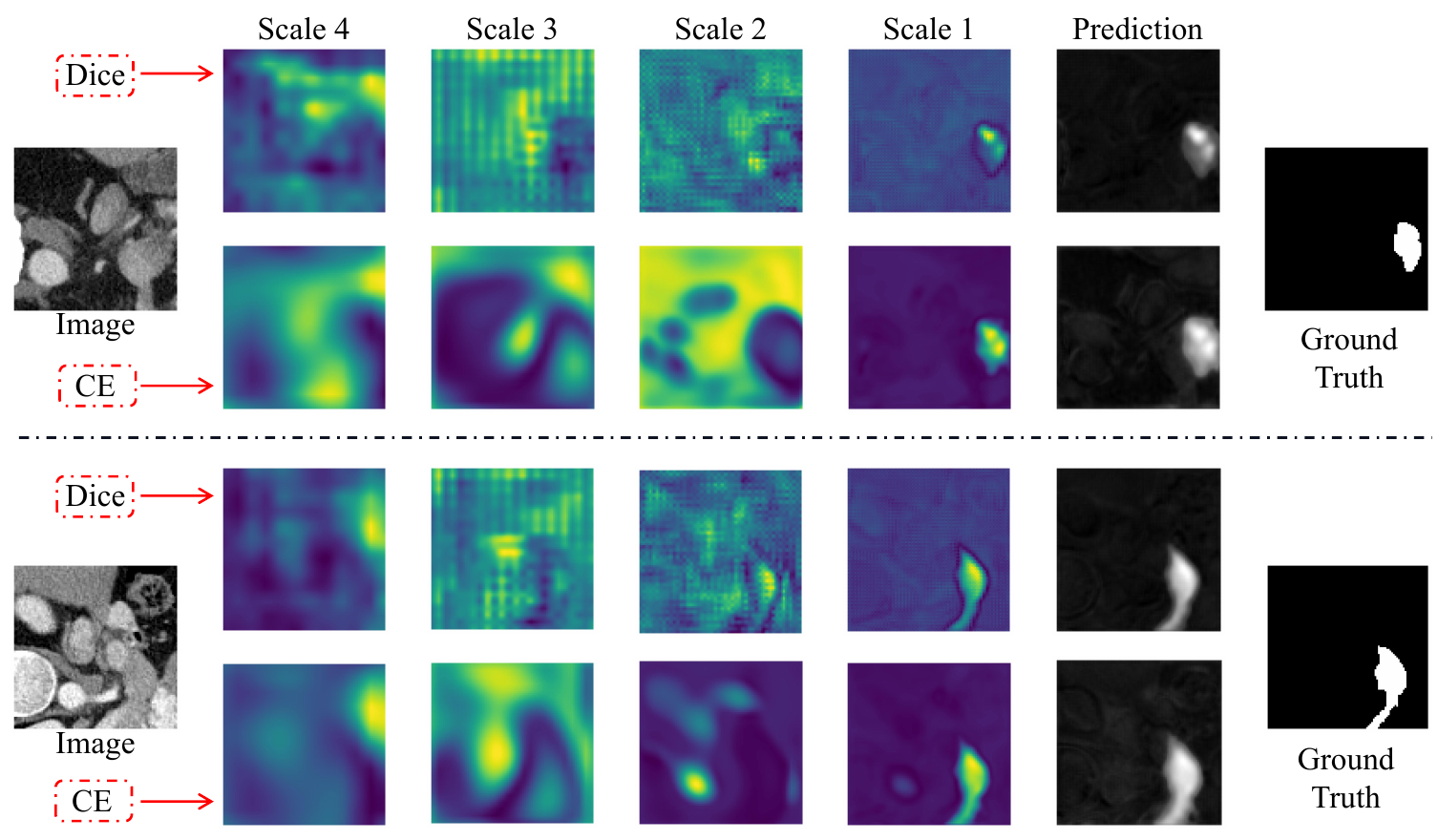}
    \caption{Decoder features obtained from different up-sampling stages. High-level to low-level features are listed from Scale4 to Scale1.}
    \label{stage_wise_fm}
\end{figure*}

\section{Results}

We compared our LeFeD with eight SOTA and relevant SSL methods: (1)~UA-MT~\cite{yu2019uncertainty} incorporates model ensemble and entropy estimation to enhance pseudo-label quality; (2)~SASSNet~\cite{li2020shape} learns from unlabeled data with shape prior; (3)~DTC~\cite{luo2021semi} investigates consistency between segmentation and regression tasks; (4)~ASE-Net~\cite{lei2022semi} uses dynamic inference to better adapt model to unseen test data; (5)~SS-Net~\cite{wu2022exploring} tries to smooth the decision boundary by imposing a strong constraint in the situation of injecting adversarial noise; (6)~MC-Net+~\cite{wu2022mutual} pays attention to the cycled pseudo-label generation and uses a sharpening function for entropy minimization; (7)~FUSSNet~\cite{xiang2022fussnet} designs uncertainty estimation schemes to split certain and uncertain regions; (8)~BCP~\cite{bai2023bidirectional} aligns the kernel distribution with bidirectional copy-paste augmentation.

\subsection{Comparisons on Pancreas Dataset }
Table~\ref{table1} presents the comparison with the other eight competitive methods. 
It is notable that LeFeD exhibits marked performance advantages over the second-best method, BCP~\cite{bai2023bidirectional}. Specifically, when operating with only 10\% (6) of the data labeled, LeFeD outperforms BCP by margins of 1.68\% and 2.27\%, as measured by the Dice and Jaccard scores, respectively. Although a marginal performance gap of 0.22\% is observed in the Dice score in favor of BCP when the labeled data is increased to 20\%, LeFeD compensates with a noteworthy improvement of 1.43\% on the 95HD score.
Beyond that, when compared with MC-Net+~\cite{wu2022mutual} that employs three distinct decoders for consistency learning, LeFeD demonstrates significant Dice performance gains of 5.51\% and 3.32\% under label percentages of 10\% and 20\%, respectively.
These results robustly validate the superior performance and inherent simplicity of our LeFeD. Remarkably, LeFeD achieves this by capitalizing on inherent inconsistencies within feature learning and eliminating the need for additional uncertainty estimation schemes or strict constraints.
Moreover, the first two rows of Fig.~\ref{segmentation_results} visually represent the segmentation outcomes of various SSL techniques. Within these illustrations, it is apparent that LeFeD consistently produces the most comprehensive and accurate segmentations, particularly in regions that are traditionally challenging for segmentation algorithms.


\begin{figure*}[t]
    \centering
    \includegraphics[width=0.9\textwidth]{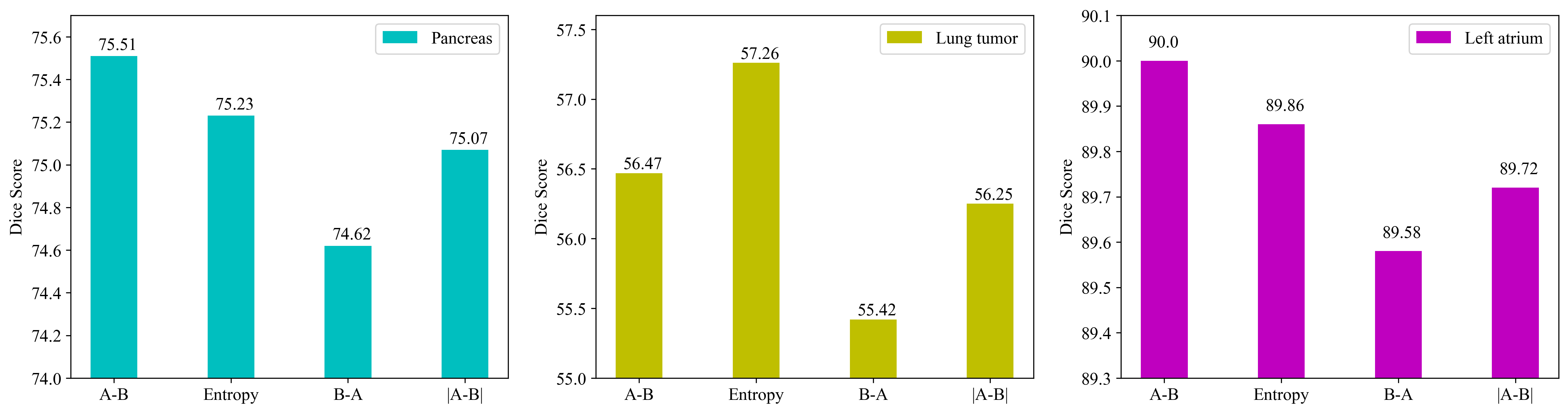}
    \caption{Discussion of the way to encode discrepancy. Four encoding types are compared, including \textbf{A-B}, \textbf{B-A}, \textbf{$|$A-B$|$} and \textbf{Entropy}. \textbf{A-B} means the discrepancy is generated by subtracting the output of decoder B from decoder A. Symbol $|\cdot|$ represents taking absolute value.}
    \label{de}
\end{figure*}

\subsection{Comparisons on Lung Tumor Dataset }
We have further extended the applicability of LeFeD to encompass the task of lung tumor segmentation. As evidenced in Table~\ref{table_l}, LeFeD consistently outperforms across all evaluation metrics under varying proportions of labeled data. For instance, when compared to the second-best performing model, BCP~\cite{bai2023bidirectional}, LeFeD exhibits performance gains of 2.11\% and 0.76\% in Dice and HD scores respectively with 20\%(10) labeled data, and 1.17\% and 0.66\% with 30\%(15) labeled data.
Furthermore, we observe a noticeable performance degradation in the task of lung tumor segmentation when compared to pancreas segmentation and left atrium segmentation. This performance dip is primarily attributable to the complicated shapes and textures of lung tumors. Encouragingly, LeFeD demonstrates a lower rate of performance degradation relative to competing models. This suggests that LeFeD holds considerable potential for effectively addressing challenging tasks, even when constrained by limited labeled data.
Additionally, the intermediate two rows of Fig.~\ref{segmentation_results} visually present the outcomes of lung tumor segmentation by LeFeD. These results reveal that LeFeD can capture more complete tumor areas with fewer false positives.


\subsection{Comparisons on Left Atrium Dataset }
As shown in Table~\ref{table2}, LeFeD exhibits a compelling capability, nearly matching the fully supervised baseline even when operating with a limited labeled dataset, constituting merely 20\% (16 samples) of the entire pool. Specifically, LeFeD achieves a Dice score of 91.44\%, which is marginally below the fully supervised baseline of 91.47\%, and a Jaccard score of 84.30\% compared to the 84.36\% attained under full supervision.
Notably, it is interesting to find that LeFeD records an ASD score of 1.39, which intriguingly surpasses the theoretical upper bound of 1.51. This phenomenon is likely attributed to the distribution of inconsistent features, which mainly localize outside the central organs, thus empowers LeFeD to capture richer information on surface characteristics.
Moreover, as the proportion of labeled data diminishes further to 10\% (8 samples), LeFeD’s performance gains become increasingly pronounced. For example, compared to FUSSNet (as cited in \cite{xiang2022fussnet}), LeFeD secures a superior lead, outperforming it by 1.13\% and 1.81\% in terms of Dice and Jaccard scores, respectively.
As for the visualization results, the last two rows of Fig.~\ref{segmentation_results} show that LeFeD consistently generates masks that are remarkably close to the ground truth, outperforming its competitor models.

\begin{table}[t]
\centering
\caption{Discussion of hyper-parameter $t$ on three datasets.}\label{hyper_t}
\setlength{\tabcolsep}{3.6mm}
\begin{tabular}{c|cccc}
\hline
\multicolumn{1}{c|}{\multirow{2}{*}{$t$}} & \multicolumn{4}{c}{Metrics on pancreas dataset~(10\% labels)}                                                                                       \\ \cline{2-5} 
\multicolumn{1}{c|}{}                   & \multicolumn{1}{c|}{Dice~$\uparrow$} & \multicolumn{1}{c|}{Jaccard~$\uparrow$} & \multicolumn{1}{c|}{ASD~$\downarrow$} & 95HD~$\downarrow$ \\ \hline
$t$=2                                     & \multicolumn{1}{c|}{74.96}         & \multicolumn{1}{c|}{60.29}            & \multicolumn{1}{c|}{3.72}    &  12.63           \\ \hline
$t$=3                                     & \multicolumn{1}{c|}{\textbf{75.51}}         & \multicolumn{1}{c|}{\textbf{61.51}}            & \multicolumn{1}{c|}{\textbf{3.44}}           &   \textbf{11.79}          \\ \hline
$t$=4                                     & \multicolumn{1}{c|}{75.28}         & \multicolumn{1}{c|}{61.30}            & \multicolumn{1}{c|}{3.57}    &   12.16          \\ \hline
\hline
& \multicolumn{4}{c}{Metric on Lung tumor dataset~(20\% labels)}                                                                                       \\ \cline{2-5} 
 \hline
$t$=2                                     & \multicolumn{1}{c|}{57.74}         & \multicolumn{1}{c|}{42.10}            & \multicolumn{1}{c|}{2.62}    &  8.86           \\ \hline
$t$=3                                     & \multicolumn{1}{c|}{\textbf{58.79}}         & \multicolumn{1}{c|}{\textbf{43.09}}            & \multicolumn{1}{c|}{\textbf{2.11}}           &   \textbf{8.12}          \\ \hline
$t$=4                                     & \multicolumn{1}{c|}{57.02}         & \multicolumn{1}{c|}{41.77}            & \multicolumn{1}{c|}{3.06}    &   9.12          \\ \hline
\hline
& \multicolumn{4}{c}{Metric on Left atrium dataset~(10\% labels)}                                                                                       \\ \cline{2-5} 
 \hline
$t$=2                                     & \multicolumn{1}{c|}{89.55}         & \multicolumn{1}{c|}{81.03}            & \multicolumn{1}{c|}{1.89}    &  7.42           \\ \hline
$t$=3                                     & \multicolumn{1}{c|}{\textbf{90.00}}         & \multicolumn{1}{c|}{\textbf{81.99}}            & \multicolumn{1}{c|}{\textbf{1.67}}           &   \textbf{6.78}          \\ \hline
$t$=4                                     & \multicolumn{1}{c|}{89.78}         & \multicolumn{1}{c|}{81.75}            & \multicolumn{1}{c|}{1.73}    &   6.97          \\ \hline

\end{tabular}
\end{table}

\begin{table}[t]
\centering
\caption{Discussion of hyper-parameter $\lambda$ on three datasets.}\label{hyper_lambda}
\setlength{\tabcolsep}{3.0mm}
\begin{tabular}{c|cccc}
\hline
\multicolumn{1}{c|}{\multirow{2}{*}{$\lambda$}} & \multicolumn{4}{c}{Metrics on pancreas dataset~(10\% labels)}                                                                                       \\ \cline{2-5} 
\multicolumn{1}{c|}{}                   & \multicolumn{1}{c|}{Dice~$\uparrow$} & \multicolumn{1}{c|}{Jaccard~$\uparrow$} & \multicolumn{1}{c|}{ASD~$\downarrow$} & 95HD~$\downarrow$ \\ \hline
$\lambda$=1e-2                                     & \multicolumn{1}{c|}{73.87}         & \multicolumn{1}{c|}{69.76}            & \multicolumn{1}{c|}{3.90}    &  13.31           \\ \hline
$\lambda$=1e-3                                     & \multicolumn{1}{c|}{\textbf{75.51}}         & \multicolumn{1}{c|}{\textbf{61.51}}            & \multicolumn{1}{c|}{\textbf{3.44}}           &   \textbf{11.79}          \\ \hline
$\lambda$=1e-4                                     & \multicolumn{1}{c|}{75.06}         & \multicolumn{1}{c|}{60.89}            & \multicolumn{1}{c|}{3.76}    &   12.72          \\ \hline
\hline
& \multicolumn{4}{c}{Metric on Lung tumor dataset~(20\% labels)}                                                                                       \\ \cline{2-5} 
 \hline
$\lambda$=1e-2                                     & \multicolumn{1}{c|}{56.98}         & \multicolumn{1}{c|}{40.72}            & \multicolumn{1}{c|}{3.16}    & 
 8.78           \\ \hline
$\lambda$=1e-3                                     & \multicolumn{1}{c|}{\textbf{58.79}}         & \multicolumn{1}{c|}{\textbf{43.09}}            & \multicolumn{1}{c|}{\textbf{2.11}}           &   \textbf{8.12}          \\ \hline
$\lambda$=1e-4                                     & \multicolumn{1}{c|}{57.82}         & \multicolumn{1}{c|}{42.20}            & \multicolumn{1}{c|}{2.66}    &   8.53          \\ \hline
\hline
& \multicolumn{4}{c}{Metric on Left atrium dataset~(10\% labels)}                                                                                       \\ \cline{2-5} 
 \hline
$\lambda$=1e-2                                     & \multicolumn{1}{c|}{89.11}         & \multicolumn{1}{c|}{80.80}            & \multicolumn{1}{c|}{2.18}    &  7.76           \\ \hline
$\lambda$=1e-3                                     & \multicolumn{1}{c|}{\textbf{90.00}}         & \multicolumn{1}{c|}{\textbf{81.99}}            & \multicolumn{1}{c|}{\textbf{1.67}}           &   \textbf{6.78}          \\ \hline
$\lambda$=1e-4                                     & \multicolumn{1}{c|}{89.56}         & \multicolumn{1}{c|}{81.61}            & \multicolumn{1}{c|}{1.96}    &   7.01          \\ \hline

\end{tabular}
\end{table}

\begin{figure*}[t]
    \centering
    \includegraphics[width=1\textwidth]{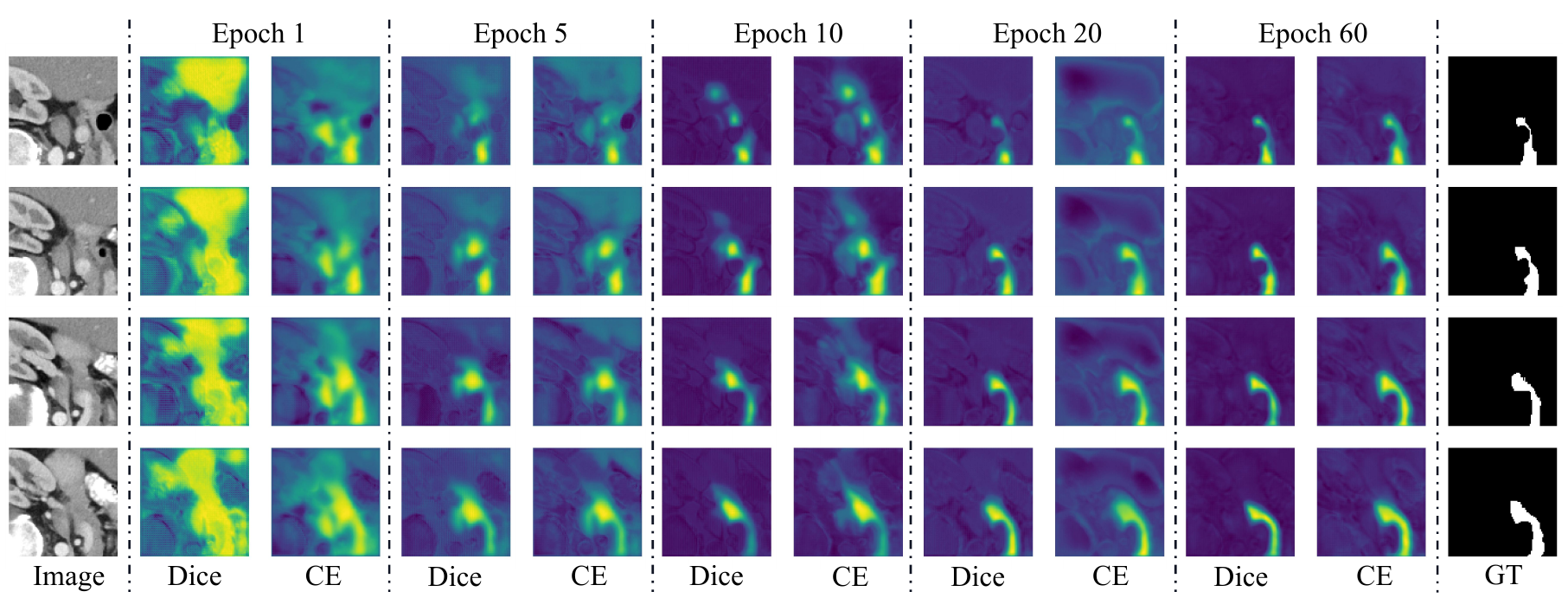}
    \caption{Visualization of the last layer decoder feature maps in different training stages, $i.e.$, from Epoch1 to Epoch60. ``Dice'' and ``CE'' represent the feature maps  produced by decoders that are optimized by Dice loss and CE loss, respectively.}
    \label{fm}
\end{figure*}

\section{Discussion}

\subsection{Ablation Study}

To evaluate the effectiveness of each component, we conduct ablation studies on the pancreas dataset under a label percentage of 10\%. As detailed in the first four rows of Table~\ref{table3}, we can observe that by solely leveraging Inconsistent Learning (IL), the performance of the LeFeD model surpasses that reported in~\cite{luo2021semi,li2020shape}, which provides empirical validation of the efficacy of employing discrepancy learning as a strategy. Besides, when different loss functions (DL) are combined with deep supervision (DS) to train the decoders, consistent gains in performance are observed across evaluations. This trend suggests that highlighting inconsistent information proves advantageous when training with unlabeled data.

We also explore the effectiveness of using the decoder features across various scales. Our results in Fig.~\ref{stage_wise_fm} show that features from scale 1 are directly aligned with the final prediction. This observation leads us to infer that the discrepancies discerned at scale 1 are potentially focused on challenging regions, such as object boundaries. Differently, features from scale 2 to scale 4 appear to contain an abundance of high-level semantic information, which somewhat complicates the interpretability of these discrepancies.
For a deeper understanding, we conduct quantitative experiments to explore whether these inconsistencies have tangible effects on learning from unlabeled data. As depicted in the last four rows of Table~\ref{table3}, discrepancies obtained from all scales have a positive influence on model performance.
We hypothesize that discrepancies derived from high-level features may be indicative of variations in the model's comprehension of image content. Therefore, when presented with identical input, the discrepancies discerned at each scale prove to be of significant consequence, particularly as the decoders endeavor to yield consistent predictions.

\subsection{Analysis of Hyper-parameters}

In our experiments, $t$ and $\lambda$ serve as two critical hyper-parameters. Specifically, $t$ denotes the iteration times for each sample in discrepancy learning, while $\lambda$ is a weighting factor to balance the influence of discrepancy. 
As evidenced in Table~\ref{hyper_t}, LeFeD exhibits optimal performance when setting $t$ to 3. When $t$ is set to a substantially larger value, the model may overfit as the discrepancy decreases gradually. Oppositely, when $t$ is set to a smaller value, the process of discrepancy learning is inadequate. 
From Table~\ref{hyper_lambda}, we can observe that the most suitable value for $\lambda$ is 1e-3. It is worth highlighting that $\lambda$ has a greater impact on results than $t$, which means controlling $\lambda$ is more important than $t$.

\subsection{Analysis of Discrepancy Encoding}
In this part, we discuss different strategies for encoding discrepancy, including $A-B$, $B-A$, $|A-B|$ and Entropy. 
$A-B$ means the discrepancy is calculated by subtracting the output of decoder B from that of decoder A. The symbol $|\cdot|$ means taking the absolute value. 
From Fig.~\ref{de}, we can see $A-B$ generally yields superior results in comparison to $B-A$, despite the performance gap being relatively narrow (commonly around a 1\% difference in Dice score).
The main reason for the phenomenon lies in the distinct supervision levels of the two decoders: decoder A is subjected to deep supervision, theoretically endowing it with enhanced segmentation capabilities relative to decoder B. Consequently, the discrepancy computed through $A-B$ is more likely to highlight the false negatives in the segmentation, whereas $B-A$ is more inclined to highlight the false positives.
Therefore, $A-B$ emerges as the more favorable configuration for our purposes. 
$|A-B|$ and Entropy are between the case of B-A and A-B, so their effects are in the medium.

\subsection{Visualization of Features in Different Training Stages}
As displayed in Fig.~\ref{fm}, we visualize the feature maps of the last decoder layers when training with different loss functions and training epochs. We can find that the features vary significantly especially in the early training stages (ranging from Epoch 1 to Epoch 10, approximately within the first 1500 iterations). It is clear to see that Dice loss pays attention to region-wise predictions whereas CE loss focuses on voxel-wise predictions, but the final training objective is the same, that is making consistent predictions. Beyond that, we can find the discrepancy acquired from the last layer typically exists around the boundary and ambiguous regions, which is beneficial to model performance if utilized. For example, according to Table~\ref{table1}, LeFeD surpasses SS-Net by 5.77\% on Dice score when using 10\% labels on the Pancreas dataset.

\subsection{Model Size Comparison}
We further conduct experiments to compare various models in terms of their parameters and multiply-accumulate operations (MACs). These results are reported using an input size of 96~$\times$96~$\times$96. It is noteworthy that, despite the observed increase in computational cost when compared to the baseline V-Net~\cite{milletari2016v}, there is a marked improvement in the results. For instance, when utilizing 10\% of the labeled data on the left atrium dataset, we observe gains of 7.26\% and 6.57\% in the Dice and HD scores, respectively.
Furthermore, in comparison to the MC-Net model~\cite{wu2021semi}, an SSL method incorporating two decoders, our LeFeD presents Dice score performance gains exceeding 3\% when trained with 10\% of the labels on three datasets, while keeping nearly the same computational cost. These observations validate both the efficiency and effectiveness of discrepancy learning.

\begin{figure}[t]
    \centering
    \includegraphics[width=0.43\textwidth]{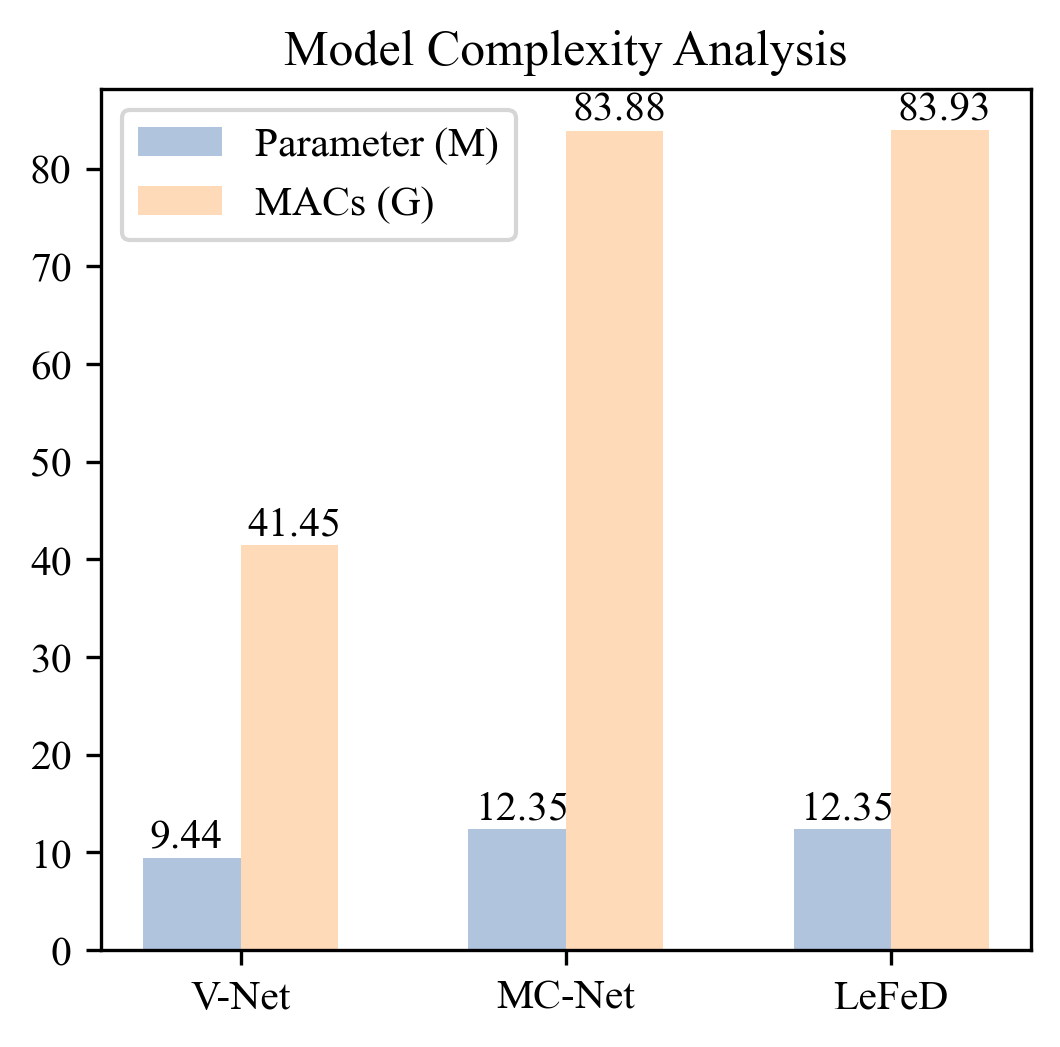}
    \caption{This figure presents the model complexity in terms of the number of parameters and multiply-accumulate operations (MACs), which are recommended by~\cite{wu2022mutual,luo2021semi}.}
    \label{mca}
\end{figure}

\section{Conclusion, Limitation and Future Work}
In this paper, we first analyze the treasure of discrepancy in learning towards consistent predictions, and thereby propose a novel SSL method LeFeD, which regards the discrepancy as a feedback signal and feeds it to the encoder. 
Different from priors that emphasize filtering out inconsistent regions or exploring diverse regularized constraints, our approach stands as intuitively effective. Specifically, it involves the training of differentiated decoders, followed by the learning process informed by the resultant discrepancy. 
Despite its apparent simplicity, LeFeD presents a new insight to learn from unlabeled data. Compared to other SSL methods, LeFeD is more generic since there is no special design to learn lesion or organ-related information. 
Empirical results further substantiate the efficacy and superiority of LeFeD, as it consistently surpasses competing methods, particularly in scenarios where annotation resources are limited.

However, it is important to note potential challenges associated with the direct application of LeFeD in multi-center and multi-domain contexts. This is attributed to the assumption in SSL settings that the training and test data are sampled from a similar distribution. Looking forward, we plan to extend LeFeD to accommodate a broader range of datasets and tasks. This direction of development aims to enhance the reliability and applicability of LeFeD, making it a feasible option for clinical deployment.

\bibliographystyle{ieeetr}
\bibliography{refs}

\end{document}